\documentclass[11pt]{article}

\usepackage[final]{acl}

\usepackage{orcidlink}
\usepackage{times}
\usepackage{latexsym}
\usepackage{booktabs}
\usepackage{multirow}
\usepackage[table]{xcolor}
\usepackage{graphicx}
\usepackage{algorithm}
\usepackage{algpseudocode}
\usepackage{amssymb}
\usepackage{amsmath}
\usepackage[T1]{fontenc}

\usepackage[utf8]{inputenc}

\usepackage{microtype}

\usepackage{inconsolata}

\usepackage{graphicx}

\title{Information-Guided Frontier Decoding: Contextual Utility-Driven Commitment in dMLLMs}

\author{
\textbf{Xingyou Fang}$^{1,*}$ \quad
\textbf{Jingxing Zhong}$^{1,*}$ \quad
\textbf{Xiaosong Yuan}$^{2,\dagger}$ \quad
\textbf{Xiaofeng Zhang}$^{3}$ \\[3pt]
$^{1}$Fuzhou University \\
$^{2}$Jilin University \\
$^{3}$Shanghai Jiao Tong University \\
{\small
\texttt{832404107@fzu.edu.cn},
\texttt{zhongjingxing083@gmail.com},
\texttt{yuanxs19@mails.jlu.edu.cn},
\texttt{SemiZxf@163.com}
}
}

\begin{document}
\maketitle

\renewcommand{\thefootnote}{\fnsymbol{footnote}}
\footnotetext[1]{These authors contributed equally to this work.}
\footnotetext[2]{Corresponding author.}

\begin{abstract}

Decoding quality in diffusion multimodal language models (dMLLMs) depends heavily on the order in which masked tokens are committed. Existing confidence-based strategies prioritize locally easy tokens, but confidence does not necessarily reflect contextual usefulness. As a result, structurally easy tokens such as punctuation may be committed before informative semantic anchors, weakening context propagation and increasing error accumulation. We propose Information-Guided Frontier Decoding (IGFD), a training-free decoding strategy that ranks candidates using token confidence, neighborhood uncertainty, and structural commitment risk. IGFD encourages early commitment of reliable semantic anchors while delaying fragile structural tokens, improving contextual support during decoding. A dynamic candidate frontier further constrains token selection to locally expandable regions under the same decoding budget. The method requires no additional training, auxiliary models, or extra forward passes. Experiments across multimodal understanding, reasoning, grounding, and hallucination benchmarks show that IGFD consistently outperforms existing decoding strategies across the majority of benchmarks and diffusion MLLM backbones under identical decoding budgets.

\end{abstract}
\begin{figure}[t]

  \centering

  \includegraphics[width=\columnwidth]{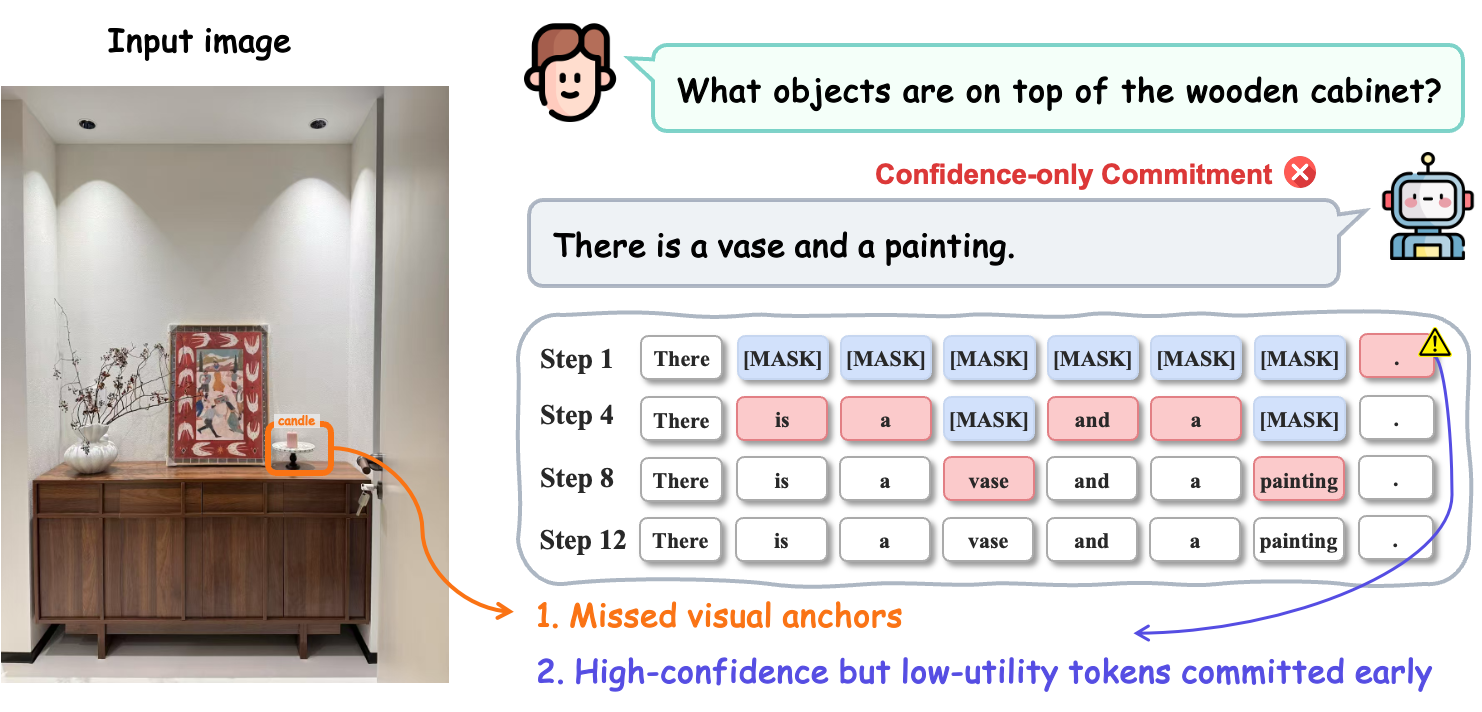}

  \caption{Confidence-only decoding may commit locally easy but low-utility tokens too early, while delaying content-bearing anchors that are more helpful for resolving nearby masked positions.}

  \label{fig:Motivation}

\end{figure}

\begin{figure*}[t]
  \centering
  \includegraphics[width=\textwidth]{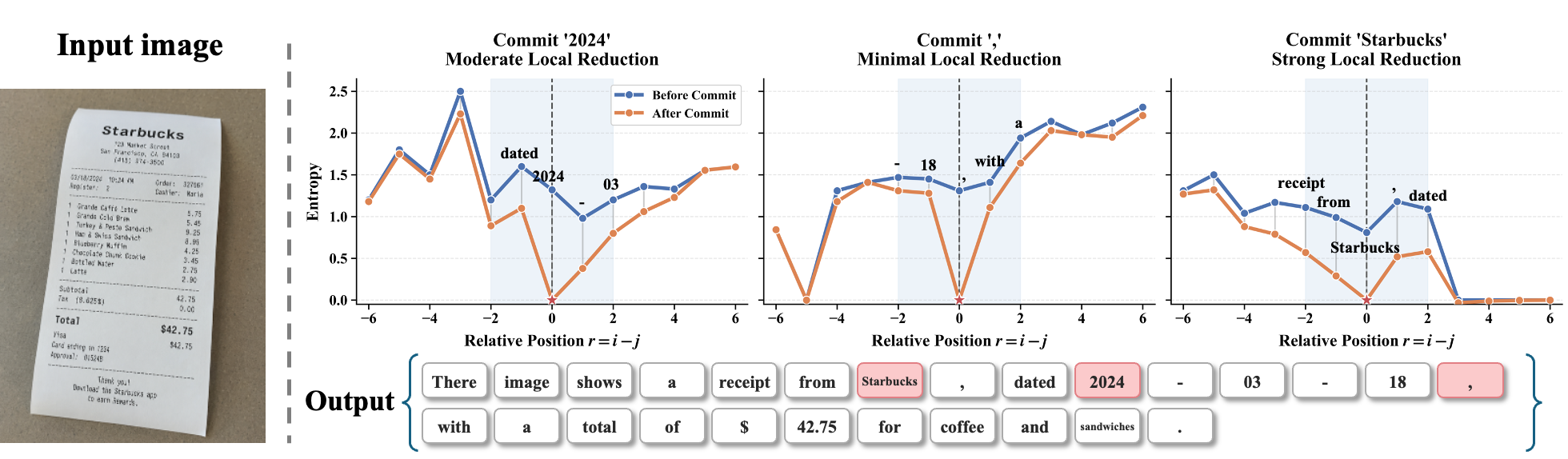}
  \caption{\textbf{Single-token intervention.} Each subplot compares local entropy profiles before and after committing exactly one token. The x-axis denotes relative position to the committed token. The semantic anchor “Starbucks” produces the largest local entropy reduction, the date token “2024” produces a moderate reduction, and punctuation produces minimal change.
    }
  \label{fig:motivation_entropy}
\end{figure*}

\section{Introduction}

Diffusion multimodal large language models (dMLLMs) \cite{du2026r,you2025llada,bie2025llada2,li2026lavida} have emerged as a promising alternative to autoregressive generation. Rather than generating tokens left to right, they begin with a masked sequence and iteratively denoise multiple positions in parallel using bidirectional context, following a broader line of masked iterative generation methods~\citep{ghazvininejad2019mask}. This flexible decoding paradigm is well suited to multimodal reasoning, long-form generation, and structured outputs such as code.

However, diffusion decoding quality depends heavily on the order in which masked tokens are committed, since finalized tokens become fixed context for later predictions. Existing strategies mainly rely on confidence \cite{cai2026confidence,lee2025lookahead}, which captures local certainty but does not necessarily measure whether a token is useful for stabilizing unresolved neighbors. As a result, confidence-based decoding may commit locally easy but low-utility tokens, such as punctuation, whitespace, or formatting symbols, before content-bearing anchors such as entities, numbers, variables, or key reasoning words \cite{fang2026locally,zhai2026core}. This produces a locally confident but contextually suboptimal commitment order, weakening the evidence available to later predictions.

We identify two commitment-order failures behind this mismatch. First, \textbf{contextual utility blindness}: confidence measures whether the model is locally certain about a token, but not whether committing that token will reduce uncertainty around it. A token with moderate confidence but highly uncertain neighbors may therefore be more valuable than a high-confidence token in an already stable region\cite{farquhar2024detecting,kuhn2023semantic,devlin2019bert}. Second, \textbf{structural commitment risk}: punctuation, whitespace, and formatting tokens are often easy to predict early, yet committing them too soon can lock unstable boundaries into the sequence. These failures cause weakly informative tokens to be committed early while useful semantic anchors are delayed, leaving nearby masked positions without sufficient local support\cite{park2025flexible,hamilton2025lost}.

To address these issues, we propose \textbf{Information-Guided Frontier Decoding} (IGFD), a training-free decoding strategy for dMLLMs. IGFD changes the commitment objective from selecting the easiest tokens to selecting tokens that are both reliable and useful. For each candidate position, IGFD computes a composite commitment score that combines token confidence, neighborhood uncertainty and structural commitment risk. The confidence-neighborhood product serves as a lightweight proxy for contextual utility, favoring reliable tokens that may help stabilize nearby uncertain positions. A structure-aware adjustment further encourages content-bearing anchors while delaying fragile structural tokens. Thus, improvements come from a better commitment order rather than increased computation or model modification.



    
    
Our contributions are as follows:

\begin{itemize}
    \item We identify contextual utility blindness and structural commitment risk as two commitment-order failures in confidence-based dMLLM decoding.

    \item We propose an information-guided commitment score that combines reliability, neighborhood utility, and structural safety.

    \item We design a dynamic candidate frontier that controls where tokens are eligible for commitment under a fixed decoding budget.
\end{itemize}

\section{Related Work}

Masked diffusion decoding \cite{austin2021structured,nie2026large,bie2025llada2,ye2025dream,zhang1,zhang2,zhang3,zhang4,zhang5,zhang6,zhang7,zhang8,zhang9} generates sequences by beginning with all positions masked. During each refinement step, the model simultaneously predicts distributions for every unresolved token, after which a sampling or selection strategy decides which subset of positions should be fixed before the next iteration.

\subsection{Ordering Effects in Masked Generative Decoding.}
Prior work has shown that the order in which tokens are committed is crucial for generation quality. Kim et al.~\citep{kim2025train} observe that order-agnostic training leads to non-uniform prediction difficulty across masked positions. They further demonstrate that prioritizing positions with higher prediction confidence can substantially improve performance over decoding schemes that follow a predetermined order. This suggests that commitment order is not merely an implementation detail, but a key factor affecting the final output quality.

Zhou et al.~\citep{zhou2026next} introduce HDLM at NeurIPS 2025, which incorporates a coarse-to-fine semantic hierarchy into the training objective. Their results indicate, from the perspective of training design, that semantically important or well-grounded positions should be resolved earlier. However, neither of these lines of work provides an inference-time decoding strategy that can dynamically preserve an appropriate commitment order under visual conditioning or long-form generation settings. 
\subsection{Structured Decoding with Localized Segments.}
Another line of work modifies the decoding process by imposing or exploiting block-level structure. Block Diffusion divides a sequence into fixed-size segments and performs autoregressive diffusion within each block \citep{arriola2025block}. AdaBlock-dLLM further adjusts block boundaries according to confidence volatility, allowing block partitions to better align with semantic changes \citep{lu2025adablock}. WavefrontDiffusion expands decoding outward from already finalized positions, thereby propagating generation from reliable anchors \citep{yang2025wavefrontdiffusion}. Deferred Commitment Decoding postpones the commitment of high-uncertainty tokens near sliding-window boundaries \citep{shu2026deferred}. AHD monitors token stability through historical convergence patterns and uses this information to enable earlier cross-block decoding \citep{zou2026breaking}. 
\subsection{Efficiency-Oriented Commitment Criteria.}
Confidence-based scoring is widely used to decide which tokens should be committed or accelerated during diffusion decoding. 
Wu et al.~\citep{wu2025fast} propose Fast-dLLM, which combines KV caching with confidence-threshold decoding to reduce inference cost. 
Wei et al.~\citep{wei2025accelerating} introduce SlowFast Sampling, a two-stage strategy that switches between exploratory and accelerated decoding according to token confidence and positional stability. 
Israel et al.~\citep{israel2506accelerating} propose Adaptive Parallel Decoding, which adjusts the sampling budget using diffusion marginals and autoregressive mixtures. 
AMOM~\citep{xiao2023amom} also improves refinement by selectively remasking uncertain positions.

\subsection{Contextual Dependence and Decoding Robustness.}
Recent studies show that token confidence alone is often insufficient for maintaining contextual consistency in long-form decoding. 
Chen et al.~\citep{chen2025beyond} propose Coherent Contextual Decoding, which measures trajectory consistency through conditional mutual information over historical context. 
Context-Aware Decoding~\citep{shi2024trusting} and C-PMI calibrated decoding~\citep{ren2023c} further adjust token probabilities by contrasting them with contextual evidence, encouraging generations to rely more strongly on the given condition. 
Li et al.~\citep{li2025diffusion} propose Prophet, which analyzes commitment risk through a two-phase decoding pattern and motivates more adaptive threshold control.

Other work focuses on diagnosing failure modes in confidence-based decoding. 
Huang et al.~\citep{huang2026empiricalanalysisdecodingbiases} identify boundary bias and trivial-token bias as two common problems caused by local confidence scoring, while Zhang et al.~\citep{zhang2023redi} propose ReDi to revise unreliable predictions through self-reflective remasking. 
Zhao et al.~\citep{zhao2026context} introduce CoTA, showing that context tokens can act as information anchors and that caching may disturb their information flow. 
Hong et al.~\citep{hong2026mitigating} identify mask-prior drift and positional-attention collapse as key sources of repetition and visual-grounding degradation in diffusion vision-language models. 
DeCoRe~\citep{gema2024decore} also improves contextual faithfulness by contrasting retrieval-head behavior with conditional-entropy signals. 
Together, these works suggest that robust decoding should account for contextual dependency, positional stability, and attention behavior rather than relying solely on isolated token confidence.

\begin{figure*}[t]
  \centering
  \includegraphics[width=\textwidth]{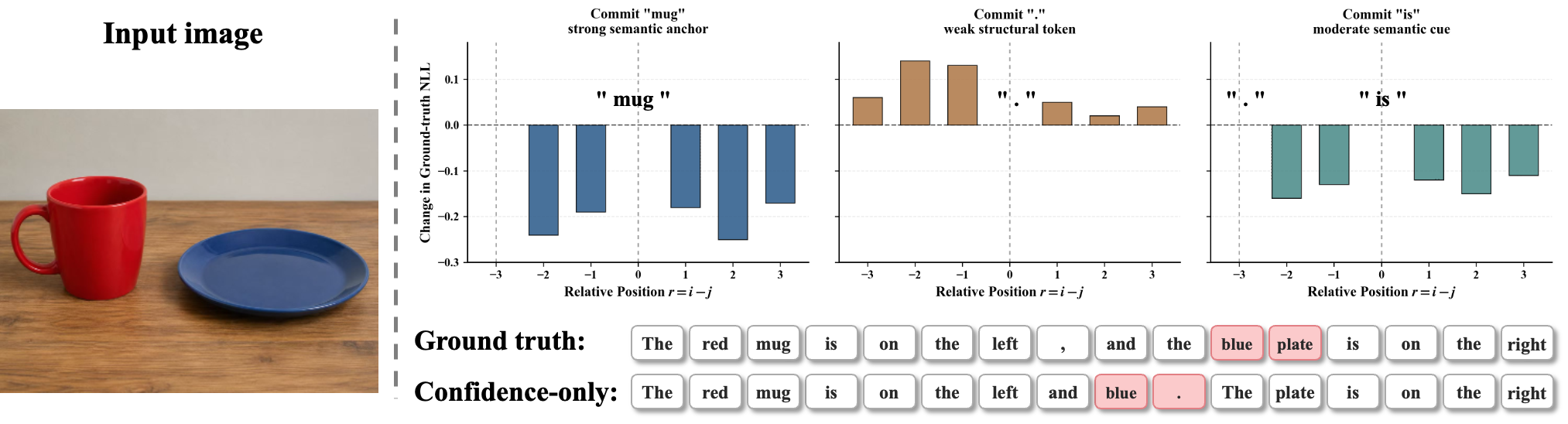}
\caption{
\textbf{Single-token intervention on nearby ground-truth NLL.}
After committing one token, we measure the change in NLL of neighboring correct tokens. Semantic tokens such as ``mug'' and ``is'' reduce local NLL, whereas the structural token ``.'' provides little benefit and may increase prediction loss.
}
  \label{fig:motivation_nll}
\end{figure*}

\section{Motivation}

Diffusion multimodal large language models decode by repeatedly predicting all masked positions and committing a subset of them. Since committed tokens become fixed context for later predictions, commitment order directly affects the uncertainty of the remaining sequence. Existing confidence-based decoders implicitly assume that high-confidence tokens are good commitments. We question this assumption: a token may be easy to predict but provide little support to its neighbors.


For each masked position $i$, the model predicts a distribution
\[
p_{t,i}(v)=p_\theta(x_i=v\mid x_t,c), \quad v\in\mathcal{V}.
\]
We measure local uncertainty using prediction entropy:
\begin{equation}
H_{t,i}
=
-\sum_{v\in\mathcal{V}}p_{t,i}(v)\log p_{t,i}(v).
\end{equation}
Lower entropy after intervention indicates that the committed token provides useful local context for nearby masked positions.

\subsection{Finding 1: Semantic Tokens Provide Unequal Local Support}

Figure~\ref{fig:motivation_entropy} compares local entropy profiles before and after committing different token types. The x-axis denotes the relative position $r=i-j$ to the committed token. The semantic anchor ``Starbucks'' produces a strong entropy reduction around the committed position, while the date token ``2024'' gives a moderate reduction. In contrast, committing punctuation produces only minimal local reduction.

This shows that different committed tokens provide very different amounts of contextual support. Semantic tokens can act as local anchors that make nearby masked positions easier to predict, whereas structural tokens may contribute little even when they are confidently predicted. Therefore, confidence alone cannot measure the contextual value of a commitment.

\subsection{Finding 2: Structural Tokens Can Be Weak or Harmful Commitments}

Entropy measures uncertainty of the model distribution, but we also want to know whether a commitment helps the model predict the correct neighboring tokens. For each nearby ground-truth token $x_i^\ast$, we compute its negative log-likelihood:
\begin{equation}
\mathrm{NLL}_{t,i}
=
-\log p_\theta(x_i^\ast \mid x_t,c).
\end{equation}
After committing token $j$, we measure the local change
\begin{equation}
\Delta \mathrm{NLL}_{i}
=
\mathrm{NLL}_{t+1,i}
-
\mathrm{NLL}_{t,i}.
\end{equation}
A negative value means the intervention makes the correct neighboring token easier to predict.

Figure~\ref{fig:motivation_nll} shows that committing semantic tokens such as ``mug'' and ``is'' reduces nearby ground-truth NLL, indicating improved prediction of correct neighboring tokens. However, committing the structural token ``.'' provides little benefit and can even increase local NLL. This suggests that punctuation and other structural tokens can be risky early commitments: they may be locally easy to predict, but they do not necessarily improve nearby semantic prediction.

\subsection{Implication}

The two findings expose a mismatch in confidence-based decoding: confidence measures local certainty, but not contextual usefulness. This leads to \textbf{contextual utility blindness}, where easy tokens are preferred over tokens that better support their neighbors. It also leads to \textbf{structural commitment risk}, where punctuation or formatting tokens are committed before the surrounding semantic content is stable. Both failures suggest that commitment order should be governed not only by local confidence, but also by the contextual effect of each committed token.

\begin{figure*}[t]
  \centering
  \includegraphics[width=\textwidth]{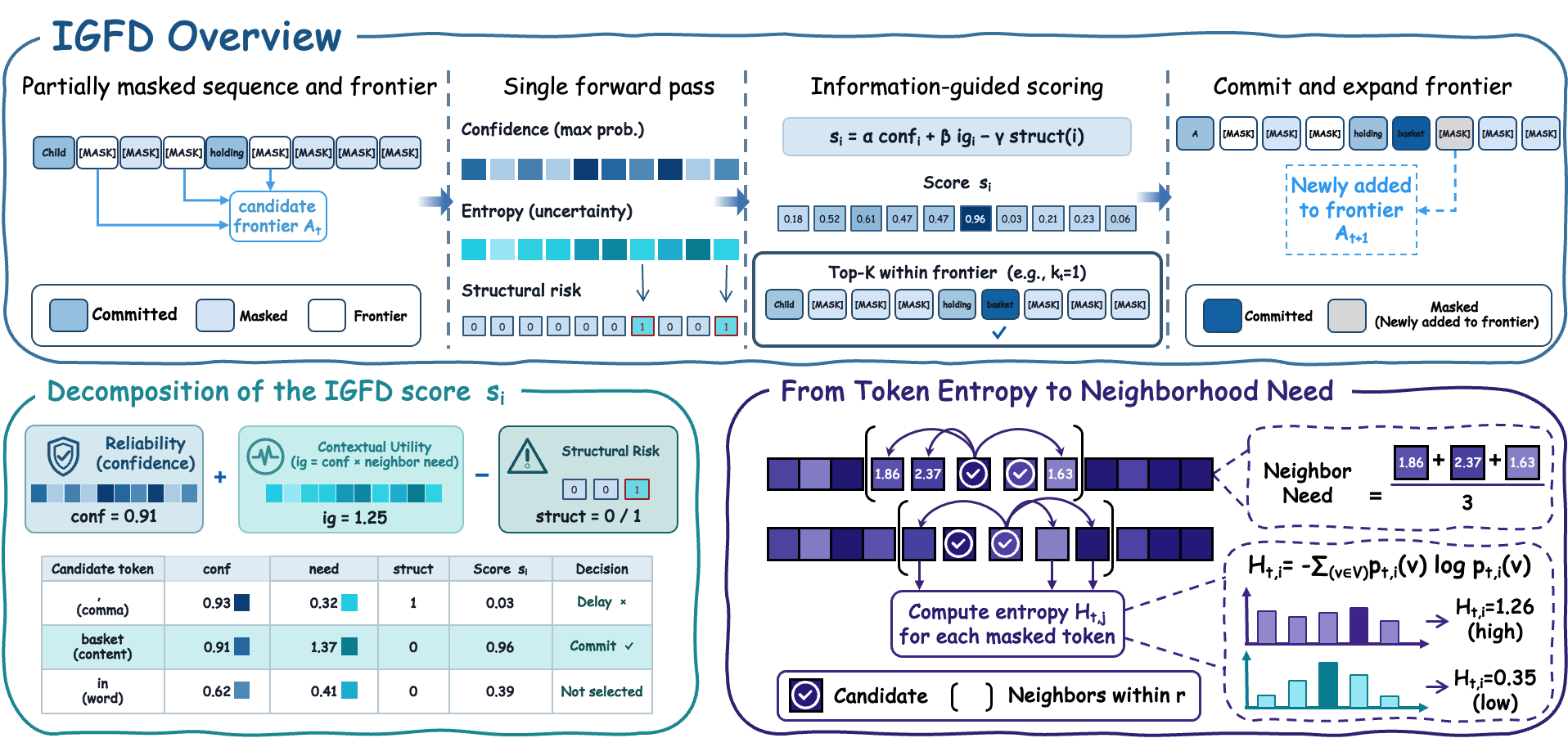}
  \caption{
  Framework of IGFD. Given a partially masked sequence, IGFD first builds a dynamic candidate frontier, then computes confidence, entropy-based neighborhood need, and structural risk for each candidate. These signals are combined into an information-guided commitment score, which selects the top-ranked tokens to commit and then expands the frontier for the next denoising step.
  }
  \label{fig:Framework}
\end{figure*}

\section{Method}

Motivated by the observations above, we propose \textbf{Information-Guided Frontier Decoding} (IGFD), a training-free decoding strategy that prioritizes masked positions according to both their prediction reliability and their potential to reduce uncertainty in nearby unresolved tokens. Unlike confidence-only decoding, IGFD avoids treating positions independently and discourages early commitment of locally easy but low-utility structural tokens.

At each denoising step, IGFD uses one model forward pass to score candidates with confidence, neighborhood uncertainty, and structural risk, then commits the top-ranked positions under the same per-step budget as the baseline.

\subsection{Preliminaries}

Let $x_t = (x_{t,1}, \ldots, x_{t,N})$ denote the partially denoised sequence at decoding step $t$, where ungenerated positions are represented by a special token $[\mathrm{MASK}]$. Let
\begin{equation}
   M_t = \{ i \mid x_{t,i} = [\mathrm{MASK}] \},
\end{equation}

be the set of masked positions. Given the current sequence $x_t$ and conditioning context $c$, the diffusion language model predicts a distribution over the vocabulary for each masked position:
\begin{equation}
p_{t,i}(v) = p_\theta(x_i = v \mid x_t, c), \quad v \in \mathcal{V}.
\end{equation}
The top-1 prediction and its confidence are
\begin{equation}
\hat{x}_{t,i} = \arg\max_{v \in \mathcal{V}} p_{t,i}(v),
\end{equation}
\begin{equation}
\mathrm{conf}_{t,i} = \max_{v \in \mathcal{V}} p_{t,i}(v).
\end{equation}

\subsection{Neighborhood Uncertainty}

Confidence measures whether the model is reliable at the current position, but it does not measure whether committing this position is useful for the surrounding context. To capture contextual utility, we estimate the uncertainty of nearby masked positions.

For each masked position $i$, we compute the entropy of the model prediction:
\begin{equation}
H_{t,i} = - \sum_{v \in \mathcal{V}} p_{t,i}(v) \log p_{t,i}(v).
\end{equation}
Higher entropy indicates that the model is less certain about the token at position $i$.

We then define the neighborhood need of position $i$ as the average entropy of masked neighbors within radius $r$:
\begin{equation}
\mathrm{need}_{t,i}
=
\frac{
\sum_{j \in \mathcal{N}_r(i) \cap M_t} H_{t,j}
}{
\max(1, |\mathcal{N}_r(i) \cap M_t|)
},
\end{equation}
where
\begin{equation}
\mathcal{N}_r(i) = \{ j \mid 0 < |j-i| \le r \}.
\end{equation}
Intuitively, $\mathrm{need}_{t,i}$ measures how uncertain the local neighborhood around position $i$ remains. A high value means that nearby masked tokens still lack sufficient contextual support.






\subsection{Information-Guided Commitment Score}

IGFD prioritizes positions that satisfy two conditions simultaneously: the model is confident about the current token, and nearby masked positions are still uncertain. We use the product of confidence and neighborhood need as a lightweight proxy for contextual utility:
\begin{equation}
\mathrm{ig}_{t,i}
=
\mathrm{conf}_{t,i}
\cdot
\mathrm{need}_{t,i}.
\end{equation}

The final commitment score is defined as
\begin{equation}
s_{t,i}
=
\alpha \cdot \mathrm{conf}_{t,i}
+
\beta \cdot \mathrm{ig}_{t,i}
-
\gamma \cdot \mathrm{struct}(\hat{x}_{t,i}),
\end{equation}
where $\alpha$, $\beta$, and $\gamma$ are scalar weights.


We implement $\mathrm{struct}(\cdot)$ as a lightweight tokenizer-level indicator:
\begin{equation}
\mathrm{struct}(y)
=
\mathbb{I}
[
y \in \mathcal{P}
\ \vee\
\mathrm{strip}(y)=\emptyset
\ \vee\
y \in \mathcal{S}
],
\end{equation}
where $\mathcal{P}$ contains punctuation symbols, $\mathrm{strip}(y)=\emptyset$ identifies whitespace-only tokens, and $\mathcal{S}$ contains special structural tokens such as newline, tab, end-of-sequence, and tokenizer-specific formatting markers. 

Thus, IGFD balances reliability, contextual utility, and structural risk, favoring informative tokens over merely easy ones and delaying fragile structural commitments until the context is more stable.

\subsection{Dynamic Candidate Frontier}

To avoid committing tokens with insufficient local context, IGFD maintains a dynamic candidate frontier $A_t \subseteq M_t$. The frontier contains masked positions that are close to already committed tokens and are therefore more likely to benefit from existing context.

Let $C_t = \{ i \mid x_{t,i} \ne [\mathrm{MASK}] \}$ denote committed positions. Given an expansion radius $R$, the candidate frontier is defined as
\begin{equation}
    A_t = \{ i \in M_t \mid \mathrm{dist}(i, C_t) \le R \}.
\end{equation}

At initialization, the frontier is seeded using the first F masked positions after the prompt, which approximates the distance-based frontier before sufficient committed tokens are available. After each commitment step, the frontier is updated by adding masked neighbors around newly committed positions. If the frontier exceeds the maximum size $F$, IGFD retains the top-$F$ positions according to the commitment score.

The frontier acts as a context availability constraint, while the information-guided score determines which candidates should be committed first.
\begin{algorithm}[t]
\caption{IGFD: Information-Guided Frontier Decoding (one round)}
\label{alg:igfd_round}
\begin{algorithmic}[1]
\Require Sequence $x$; frontier $A$; model $p_\theta$; budget $k_t$; weights $\alpha,\beta,\gamma$
\State $\ell \leftarrow p_\theta(x)$
\Comment{one forward pass}
\State Compute $p_i$, $\hat{x}_i$, $\mathrm{conf}_i$, and $H_i$ for each masked position $i$
\State $\mathrm{need}_i \leftarrow \mathrm{AvgEntropy}(\mathcal{N}_r(i) \cap M)$
\State $\mathrm{ig}_i \leftarrow \mathrm{conf}_i \cdot \mathrm{need}_i$
\State $s_i \leftarrow
\alpha \cdot \mathrm{conf}_i
+\beta \cdot \mathrm{ig}_i
-\gamma \cdot \mathrm{struct}(\hat{x}_i)$
\State $S \leftarrow \mathrm{TopK}_{i \in A \cap M}(s_i, k_t)$
\If{$|S| < k_t$}
    \State Add highest-scoring positions from $M \setminus S$ to $S$
\EndIf
\State Commit each $i \in S$ with $\hat{x}_i$
\State Expand frontier around committed positions and prune it to size $F$
\State \Return updated sequence $x$ and frontier $A$
\end{algorithmic}
\end{algorithm}
\subsection{Decoding Algorithm}

At each step $t$, IGFD commits a fixed number of positions $k_t$:
\begin{equation}
    k_t = \left\lfloor \frac{N}{T} \right\rfloor + \mathbb{I}[t \le N \bmod T],
\end{equation}
where $T$ is the total decoding budget.

The selected positions are
\begin{equation}
    S_t = \mathrm{TopK}_{i \in A_t}(s_{t,i}, k_t).
\end{equation}
For each $i \in S_t$, IGFD replaces $[\mathrm{MASK}]$ with the top-1 prediction:
\begin{equation}
    x_{t+1,i} = \hat{x}_{t,i}.
\end{equation}
All unselected masked positions remain unchanged.

If the current frontier contains fewer than $k_t$ positions, IGFD falls back to the highest-scoring positions in $M_t$ to maintain the decoding budget.

\begin{table*}[t]
\centering
\small
\setlength{\tabcolsep}{4.5pt}
\renewcommand{\arraystretch}{1.22}
\caption{Comparison of different methods across multiple benchmarks.}
\label{tab:main_results_infowave_perturbed_large}
\resizebox{\textwidth}{!}{
\begin{tabular}{llcccccccccccc}
\toprule
\multirow{2}{*}{Model} & \multirow{2}{*}{Method}
& \multicolumn{4}{c}{LLaVA-Bench}
& \multicolumn{3}{c}{CHAIR}
& \multicolumn{1}{c}{MathVista}
& \multicolumn{1}{c}{ScienceQA}
& \multicolumn{2}{c}{MME}
& \multicolumn{1}{c}{GQA} \\
\cmidrule(lr){3-6}
\cmidrule(lr){7-9}
\cmidrule(lr){10-10}
\cmidrule(lr){11-11}
\cmidrule(lr){12-13}
\cmidrule(lr){14-14}
& & all$\uparrow$ & conv$\uparrow$ & detail$\uparrow$ & complex$\uparrow$
& $C_S\downarrow$ & $C_i\downarrow$ & recall$\uparrow$
& acc$\uparrow$ & acc$\uparrow$
& cog.$\uparrow$ & perc.$\uparrow$
& EM$\uparrow$ \\
\midrule

\multirow{4}{*}{LLaDA-V}
& Original      & 50.4 & 43.4 & 57.5 & 63.6 & 5.6 & \underline{4.0} & 36.6 & 30.8 & 79.2 & 338.6 & 1411.4 & 52.3 \\
& AdaBlock      & \underline{71.8} & 59.7 & 68.8 & \underline{79.2}
                & 5.8 & 4.3 & \cellcolor{blue!8}\textbf{37.3}
                & 32.3 & 81.2
                & 350.8 & \underline{1412.6}
                & \underline{52.6} \\
& Wavefront     & 71.6 & \underline{61.6} & \underline{71.7} & 75.9
                & \underline{5.3} & 5.2 & 34.2
                & \underline{33.4} & \underline{81.8}
                & \underline{355.7} & 1397.5
                & 51.7 \\
\rowcolor{blue!10}
& \textbf{IGFD}
                & \textbf{72.8} & \textbf{63.1} & \textbf{74.6} & \textbf{80.8}
                & \textbf{5.2} & \textbf{3.7} & \underline{37.1}
                & \textbf{34.2} & \textbf{82.7}
                & \textbf{363.6} & \textbf{1422.7}
                & \textbf{53.7} \\
\midrule

\multirow{4}{*}{MMaDA}
& Original      & 35.1 & 34.4 & 39.7 & 33.7
                & 24.6 & 11.3 & 39.5
                & 23.5 & 50.2
                & 232.6 & \underline{843.9}
                & 39.4 \\
& AdaBlock      & \underline{45.9} & \underline{38.9} & 43.1 & \underline{39.9}
                & 22.8 & 11.3 & \cellcolor{blue!10}\textbf{41.8}
                & \underline{24.3} & 54.6
                & 225.8 & \cellcolor{blue!10}\textbf{845.5}
                & 42.4 \\
& Wavefront     & 44.3 & 37.2 & \underline{49.4} & 38.5
                & \underline{19.3} & \underline{10.2} & 39.2
                & 23.4 & \underline{56.3}
                & \underline{244.7} & 752.8
                & \underline{43.6} \\
\rowcolor{blue!10}
& \textbf{IGFD}
                & \textbf{47.3} & \textbf{41.9} & \textbf{50.6} & \textbf{43.7}
                & \textbf{18.7} & \textbf{9.8} & \underline{40.6}
                & \textbf{25.9} & \textbf{57.7}
                & \textbf{248.2} & 841.6
                & \textbf{44.9} \\
\midrule

\multirow{4}{*}{LaViDa}
& Original      & 51.1 & 32.8 & 57.2 & 60.1
                & 6.8 & 9.8 & \underline{34.8}
                & 41.6 & 71.2
                & 361.7 & 1348.5
                & 56.2 \\
& AdaBlock      & 75.6 & 56.8 & 73.4 & 69.3
                & 6.5 & \underline{5.3} & 33.9
                & 44.9 & \underline{72.9}
                & 368.2 & 1350.9
                & \underline{56.4} \\
& Wavefront     & \underline{77.1} & \underline{61.3} & \underline{76.3} & \underline{72.8}
                & \underline{5.7} & \cellcolor{blue!10}\textbf{4.8} & 33.6
                & \underline{45.1} & 72.1
                & \underline{372.4} & \underline{1358.1}
                & 55.9 \\
\rowcolor{blue!10}
& \textbf{IGFD}
                & \textbf{78.7} & \textbf{65.1} & \textbf{78.6} & \textbf{79.7}
                & \textbf{5.1} & 5.7 & \textbf{36.1}
                & \textbf{45.3} & \textbf{73.5}
                & \textbf{382.9} & \textbf{1366.2}
                & \textbf{58.2} \\

\bottomrule
\end{tabular}
}
\end{table*}

\section{Experiments}

We conduct experiments to evaluate whether IGFD improves decoding quality for diffusion multimodal large language models (dMLLMs). Following prior diffusion decoding work, we focus on three research questions:

\textbf{RQ1: Overall performance.} Does IGFD improve generation quality across multimodal generation, hallucination, reasoning, perception, and grounding benchmarks?

\textbf{RQ2: Component effectiveness.} How much does each part of the information-guided commitment score contribute?

\textbf{RQ3: Semantic quality and commitment behavior.} Does IGFD produce semantically more faithful outputs and more desirable commitment orders?

\subsection{Experimental Setup}

\paragraph{Models.}
We evaluate IGFD on three representative dMLLMs: LLaDA-V\cite{you2025llada}, MMaDA\cite{yang2026mmada}, and LaViDa\cite{li2026lavida}. All methods use the same backbone model within each comparison, ensuring that performance differences come only from the decoding strategy.

\paragraph{Benchmarks.}
We evaluate on six benchmarks covering complementary capabilities. LLaVA-Bench\cite{liu2023visual} measures multimodal response quality, CHAIR\cite{rohrbach2018object} evaluates object hallucination, MathVista\cite{lu2024mathvista} and ScienceQA \cite{lu2022learn} assess multimodal reasoning, MME\cite{fu2026mme} measures cognition and perception, and GQA\cite{hudson2019gqa} evaluates visual question answering accuracy.

\paragraph{Baselines.}
We compare IGFD with three decoding baselines: Original decoding, AdaBlock\cite{lu2025adablock}, and Wavefront\cite{yang2025wavefrontdiffusion}. To ensure a fair comparison, we evaluate IGFD and all baselines under identical model configurations and decoding settings.

\paragraph{Implementation Details.}
Unless otherwise specified, all methods use deterministic decoding with temperature set to 0. IGFD uses the commitment score
\[
\begin{aligned}
s_{t,i}
=&\ \alpha \cdot\mathrm{conf}_{t,i}
 + \beta \cdot \mathrm{ig}_{t,i}
 - \gamma\cdot \mathrm{struct}(\hat{x}_{t,i}).
\end{aligned}
\]
We set $\alpha=0.7$, $\beta=0.5$ and $\gamma=0.2$ by default. The neighborhood radius is set to $r=2$.




\begin{figure*}[t]
  \centering
  \includegraphics[width=\textwidth]{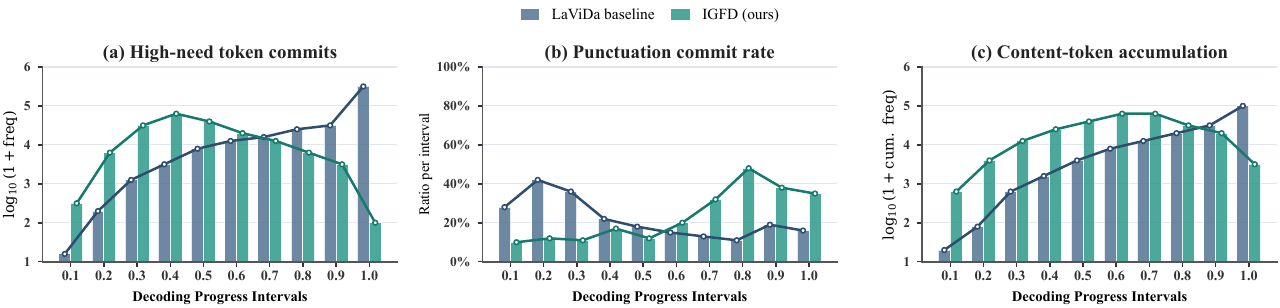}
  \caption{\textbf{Analysis of decoding behavior and commitment dynamics.} Metrics are tracked across normalized decoding progress intervals. (a) illustrates the frequency of high-need tokens (top 20\% uncertainty); (b) shows the punctuation commitment rate per interval; (c) displays the cumulative frequency of committed content-bearing tokens. IGFD significantly prioritizes critical content over structural tokens in early decoding stages.}
  \label{fig:Commitment}
\end{figure*}

\subsection{Main Results}
\label{sec:main_results}
To answer RQ1, we evaluate IGFD across three dMLLM backbones and multiple benchmarks. Table~\ref{tab:main_results_infowave_perturbed_large} reports results across three dMLLM backbones and multiple benchmarks. Overall, IGFD achieves the best performance on most metrics, consistently outperforming Original decoding, AdaBlock, and Wavefront.

For LLaDA-V, IGFD achieves the best results on nearly all benchmarks while reducing CHAIR hallucination scores. For MMaDA, IGFD improves most metrics, particularly on reasoning and grounding benchmarks, although AdaBlock performs slightly better on CHAIR recall and MME perception. For LaViDa, IGFD also achieves the best performance on most benchmarks, while Wavefront attains the lowest CHAIR $C_i$.

These results demonstrate that IGFD generalizes effectively across different dMLLM backbones and consistently improves multimodal reasoning, visual understanding, and grounding reliability.



\subsection{Ablation Study}
\label{sec:ablation}
To answer RQ2, we ablate each component of the commitment score. Table~\ref{tab:chair_ablation} reports the CHAIR ablation averaged over three dMLLMs. Removing neighborhood need increases hallucination and lowers recall, showing that contextual utility helps select more informative tokens. Removing the structural penalty also degrades CHAIR performance, confirming the importance of delaying premature punctuation and formatting tokens. Full IGFD achieves the best overall results. Detailed per-model ablations are provided in Appendix~\ref{app:ablation_details}.


\begin{table}[t]
\centering
\small
\renewcommand{\arraystretch}{1.15}
\caption{Ablation study of IGFD on CHAIR, averaged over three dMLLMs.}
\label{tab:chair_ablation}
\begin{tabular*}{0.95\linewidth}{@{\hspace{6pt}\extracolsep{\fill}}lccc@{\hspace{6pt}}}
\toprule
Variant
& $C_S \downarrow$
& $C_i \downarrow$
& Recall$\uparrow$ \\
\midrule
Confidence only
& 12.3 & 8.4 & 37.0 \\
w/o neighborhood need
& 10.3 & 6.7 & 37.5 \\
w/o structural penalty
& 10.3 & 6.7 & 37.6 \\
w/o dynamic frontier
& 10.2 & 6.6 & 37.7 \\
\textbf{Full IGFD}
& \textbf{9.7} & \textbf{6.4} & \textbf{37.9} \\
\bottomrule
\end{tabular*}
\end{table}

\subsection{Commitment Behavior Analysis}
\label{sec:commitment_behavior}
To answer RQ3 from the perspective of decoding behavior, we analyze token commitment dynamics. Figure~\ref{fig:Commitment} visualizes step-by-step commitment behaviors across normalized decoding progress. Figure~\ref{fig:Commitment}(a) shows that high-need tokens are committed substantially earlier under IGFD than under the flat baseline, indicating effective prioritization of uncertainty-reducing tokens. Figure~\ref{fig:Commitment}(b) further demonstrates that IGFD delays premature punctuation commitments until later decoding stages, preventing early structural locking. As a result, content-bearing tokens accumulate more rapidly in the early phase, as reflected by the cumulative curve in Figure~\ref{fig:Commitment}(c). Importantly, this global reordering introduces no additional forward passes and maintains the same decoding cost as Wavefront.

\begin{table}[t]
\centering
\small
\renewcommand{\arraystretch}{1.15}
\caption{Semantic quality evaluation on WikiText using BERTScore. Higher scores indicate better semantic fidelity to reference text.}
\label{tab:semantic_quality}
\begin{tabular*}{0.95\linewidth}{@{\hspace{6pt}\extracolsep{\fill}}lccc@{\hspace{6pt}}}
\toprule
Method
& Precision $\uparrow$
& Recall $\uparrow$
& F1 $\uparrow$ \\
\midrule
Original
& 0.842 & 0.834 & 0.838 \\
AdaBlock
& 0.856 & 0.849 & 0.852 \\
Wavefront
& 0.861 & 0.854 & 0.857 \\
\textbf{IGFD}
& \textbf{0.873} & \textbf{0.866} & \textbf{0.869} \\
\bottomrule
\end{tabular*}
\end{table}

\subsection{Semantic Quality Evaluation}
\label{sec:semantic_quality}
To further answer RQ3 from the perspective of semantic quality, we evaluate BERTScore\cite{zhang2019bertscore} on WikiText\cite{merity2016pointer}. We randomly sample 1,000 test sentences and generate outputs under the same decoding budget for all methods.

Table~\ref{tab:semantic_quality} shows that IGFD achieves the best BERTScore precision, recall, and F1, indicating stronger semantic alignment with the reference text. Compared with confidence-based and structured decoding baselines, IGFD better preserves semantic fidelity by prioritizing reliable content tokens and delaying fragile structural tokens until the context becomes more stable.


\section{Conclusion}

We presented IGFD, a training-free decoding strategy for diffusion multimodal large language models. IGFD addresses two commitment-order failures in confidence-based decoding: contextual utility blindness and premature structural commitment. By combining token confidence, neighborhood uncertainty and punctuation risk, IGFD prioritizes tokens that are not only reliable but also contextually informative and useful for stabilizing nearby masked positions. Experiments across multimodal generation, hallucination, reasoning, perception, and grounding benchmarks show consistent improvements over existing decoding strategies without additional training or extra model calls.

\section*{Limitations}
\label{app:limitations}
Although IGFD improves commitment ordering without additional model calls, it still has several limitations.
First, IGFD uses local neighborhood entropy as a proxy for contextual utility, which may not fully capture long-range dependencies or global discourse constraints. Second, the structural penalty relies on tokenizer-level rules. Since tokenizers represent punctuation, whitespace, and formatting tokens differently, the structural token set may require minor adaptation across model families.

\bibliography{custom}

@article{nie2026large,
  title={Large language diffusion models},
  author={Nie, Shen and Zhu, Fengqi and You, Zebin and Zhang, Xiaolu and Ou, Jingyang and Hu, Jun and Zhou, Jun and Lin, Yankai and Wen, Ji-Rong and Li, Chongxuan},
  journal={Advances in Neural Information Processing Systems},
  volume={38},
  pages={50608--50646},
  year={2026}
}

@article{bie2025llada2,
  title={Llada2. 0: Scaling up diffusion language models to 100b},
  author={Bie, Tiwei and Cao, Maosong and Chen, Kun and Du, Lun and Gong, Mingliang and Gong, Zhuochen and Gu, Yanmei and Hu, Jiaqi and Huang, Zenan and Lan, Zhenzhong and others},
  journal={arXiv preprint arXiv:2512.15745},
  year={2025}
}

@article{liu2023visual,
  title={Visual instruction tuning},
  author={Liu, Haotian and Li, Chunyuan and Wu, Qingyang and Lee, Yong Jae},
  journal={Advances in neural information processing systems},
  volume={36},
  pages={34892--34916},
  year={2023}
}

@article{park2025flexible,
  title={Flexible and efficient grammar-constrained decoding},
  author={Park, Kanghee and Zhou, Timothy and D'Antoni, Loris},
  journal={arXiv preprint arXiv:2502.05111},
  year={2025}
}

@article{kuhn2023semantic,
  title={Semantic uncertainty: Linguistic invariances for uncertainty estimation in natural language generation},
  author={Kuhn, Lorenz and Gal, Yarin and Farquhar, Sebastian},
  journal={arXiv preprint arXiv:2302.09664},
  year={2023}
}

@article{farquhar2024detecting,
  title={Detecting hallucinations in large language models using semantic entropy},
  author={Farquhar, Sebastian and Kossen, Jannik and Kuhn, Lorenz and Gal, Yarin},
  journal={Nature},
  volume={630},
  number={8017},
  pages={625--630},
  year={2024},
  publisher={Nature Publishing Group UK London}
}

@inproceedings{devlin2019bert,
  title={Bert: Pre-training of deep bidirectional transformers for language understanding},
  author={Devlin, Jacob and Chang, Ming-Wei and Lee, Kenton and Toutanova, Kristina},
  booktitle={Proceedings of the 2019 conference of the North American chapter of the association for computational linguistics: human language technologies, volume 1 (long and short papers)},
  pages={4171--4186},
  year={2019}
}

@article{zhang2019bertscore,
  title={Bertscore: Evaluating text generation with bert},
  author={Zhang, Tianyi and Kishore, Varsha and Wu, Felix and Weinberger, Kilian Q and Artzi, Yoav},
  journal={arXiv preprint arXiv:1904.09675},
  year={2019}
}

@article{merity2016pointer,
  title={Pointer sentinel mixture models},
  author={Merity, Stephen and Xiong, Caiming and Bradbury, James and Socher, Richard},
  journal={arXiv preprint arXiv:1609.07843},
  year={2016}
}

@article{fu2026mme,
  title={Mme: A comprehensive evaluation benchmark for multimodal large language models},
  author={Fu, Chaoyou and Chen, Peixian and Shen, Yunhang and Qin, Yulei and Zhang, Mengdan and Lin, Xu and Yang, Jinrui and Zheng, Xiawu and Li, Ke and Sun, Xing and others},
  journal={Advances in Neural Information Processing Systems},
  volume={38},
  year={2026}
}

@inproceedings{hudson2019gqa,
  title={Gqa: A new dataset for real-world visual reasoning and compositional question answering},
  author={Hudson, Drew A and Manning, Christopher D},
  booktitle={Proceedings of the IEEE/CVF conference on computer vision and pattern recognition},
  pages={6700--6709},
  year={2019}
}

@article{lu2022learn,
  title={Learn to explain: Multimodal reasoning via thought chains for science question answering},
  author={Lu, Pan and Mishra, Swaroop and Xia, Tanglin and Qiu, Liang and Chang, Kai-Wei and Zhu, Song-Chun and Tafjord, Oyvind and Clark, Peter and Kalyan, Ashwin},
  journal={Advances in neural information processing systems},
  volume={35},
  pages={2507--2521},
  year={2022}
}

@inproceedings{lu2024mathvista,
  title={Mathvista: Evaluating mathematical reasoning of foundation models in visual contexts},
  author={Lu, Pan and Bansal, Hritik and Xia, Tony and Liu, Jiacheng and Li, Chunyuan and Hajishirzi, Hannaneh and Cheng, Hao and Chang, Kai-Wei and Galley, Michel and Gao, Jianfeng},
  booktitle={International Conference on Learning Representations},
  volume={2024},
  pages={23439--23554},
  year={2024}
}

@inproceedings{rohrbach2018object,
  title={Object hallucination in image captioning},
  author={Rohrbach, Anna and Hendricks, Lisa Anne and Burns, Kaylee and Darrell, Trevor and Saenko, Kate},
  booktitle={Proceedings of the 2018 Conference on Empirical Methods in Natural Language Processing},
  pages={4035--4045},
  year={2018}
}

@article{ye2025dream,
  title={Dream 7b: Diffusion large language models},
  author={Ye, Jiacheng and Xie, Zhihui and Zheng, Lin and Gao, Jiahui and Wu, Zirui and Jiang, Xin and Li, Zhenguo and Kong, Lingpeng},
  journal={arXiv preprint arXiv:2508.15487},
  year={2025}
}

@inproceedings{arriola2025block,
  title={Block diffusion: Interpolating between autoregressive and diffusion language models},
  author={Arriola, Marianne and Gokaslan, Aaron and Chiu, Justin and Yang, Zhihan and Qi, Zhixuan and Han, Jiaqi and Sahoo, Subham and Kuleshov, Volodymyr},
  booktitle={International Conference on Learning Representations},
  volume={2025},
  pages={50726--50753},
  year={2025}
}

@article{yang2025wavefrontdiffusion,
  title={WavefrontDiffusion: Dynamic Decoding Schedule for Improved Reasoning},
  author={Yang, Haojin and Hu, Rui and Sun, Zequn and Zhou, Rui and Cai, Yujun and Wang, Yiwei},
  journal={arXiv preprint arXiv:2511.19473},
  year={2025}
}

@article{zhou2026next,
  title={Next semantic scale prediction via hierarchical diffusion language models},
  author={Zhou, Cai and Wang, Chenyu and Zhang, Dinghuai and Tong, Shangyuan and Wang, Yifei and Bates, Stephen and Jaakkola, Tommi},
  journal={Advances in Neural Information Processing Systems},
  volume={38},
  pages={41496--41531},
  year={2026}
}

@article{israel2506accelerating,
  title={Accelerating diffusion llms via adaptive parallel decoding},
  author={Israel, Daniel and Van den Broeck, Guy and Grover, Aditya},
  journal={Advances in neural information processing systems},
  volume={38},
  pages={52870--52888},
  year={2026}
}

@article{kim2025train,
  title={Train for the worst, plan for the best: Understanding token ordering in masked diffusions},
  author={Kim, Jaeyeon and Shah, Kulin and Kontonis, Vasilis and Kakade, Sham and Chen, Sitan},
  journal={arXiv preprint arXiv:2502.06768},
  year={2025}
}

@article{yang2026mmada,
  title={Mmada: Multimodal large diffusion language models},
  author={Yang, Ling and Tian, Ye and Li, Bowen and Zhang, Xinchen and Shen, Ke and Tong, Yunhai and Wang, Mengdi},
  journal={Advances in Neural Information Processing Systems},
  volume={38},
  pages={138867--138907},
  year={2026}
}

@article{shu2026deferred,
  title={Deferred Commitment Decoding for Diffusion Language Models with Confidence-Aware Sliding Windows},
  author={Shu, Yingte and Tian, Yuchuan and Xu, Chao and Wang, Yunhe and Chen, Hanting},
  journal={arXiv preprint arXiv:2601.02076},
  year={2026}
}

@article{zou2026breaking,
  title={Breaking Block Boundaries: Anchor-based History-stable Decoding for Diffusion Large Language Models},
  author={Zou, Shun and Wang, Yong and Chen, Zehui and Chen, Lin and Tao, Chongyang and Zhao, Feng and Chu, Xiangxiang},
  journal={arXiv preprint arXiv:2604.08964},
  year={2026}
}

@article{wu2025fast,
  title={Fast-dllm: Training-free acceleration of diffusion llm by enabling kv cache and parallel decoding},
  author={Wu, Chengyue and Zhang, Hao and Xue, Shuchen and Liu, Zhijian and Diao, Shizhe and Zhu, Ligeng and Luo, Ping and Han, Song and Xie, Enze},
  journal={arXiv preprint arXiv:2505.22618},
  year={2025}
}

@article{wei2025accelerating,
  title={Accelerating diffusion large language models with slowfast sampling: The three golden principles},
  author={Wei, Qingyan and Zhang, Yaojie and Liu, Zhiyuan and Zeng, Puyu and Wang, Yuxuan and Qi, Biqing and Liu, Dongrui and Zhang, Linfeng},
  journal={arXiv preprint arXiv:2506.10848},
  year={2025}
}

@inproceedings{xiao2023amom,
  title={Amom: adaptive masking over masking for conditional masked language model},
  author={Xiao, Yisheng and Xu, Ruiyang and Wu, Lijun and Li, Juntao and Qin, Tao and Liu, Tie-Yan and Zhang, Min},
  booktitle={Proceedings of the AAAI Conference on Artificial Intelligence},
  volume={37},
  number={11},
  pages={13789--13797},
  year={2023}
}

@inproceedings{shi2024trusting,
  title={Trusting your evidence: Hallucinate less with context-aware decoding},
  author={Shi, Weijia and Han, Xiaochuang and Lewis, Mike and Tsvetkov, Yulia and Zettlemoyer, Luke and Yih, Wen-tau},
  booktitle={Proceedings of the 2024 Conference of the North American Chapter of the Association for Computational Linguistics: Human Language Technologies (Volume 2: Short Papers)},
  pages={783--791},
  year={2024}
}

@article{gema2024decore,
  title={DeCoRe: decoding by contrasting retrieval heads to mitigate hallucinations},
  author={Gema, Aryo Pradipta and Jin, Chen and Abdulaal, Ahmed and Diethe, Tom and Teare, Philip and Alex, Beatrice and Minervini, Pasquale and Saseendran, Amrutha},
  journal={arXiv preprint arXiv:2410.18860},
  volume={10},
  year={2024}
}

@inproceedings{zhang2023redi,
  title={ReDi: efficient learning-free diffusion inference via trajectory retrieval},
  author={Zhang, Kexun and Yang, Xianjun and Wang, William Yang and Li, Lei},
  booktitle={International Conference on Machine Learning},
  pages={41770--41785},
  year={2023},
  organization={PMLR}
}

@inproceedings{ren2023c,
  title={C-PMI: Conditional pointwise mutual information for turn-level dialogue evaluation},
  author={Ren, Liliang and Sidhu, Mankeerat and Zeng, Qi and Reddy, Revanth Gangi and Ji, Heng and Zhai, ChengXiang},
  booktitle={Proceedings of the Third DialDoc Workshop on Document-grounded Dialogue and Conversational Question Answering},
  pages={80--85},
  year={2023}
}

@article{lee2025lookahead,
  title={Lookahead unmasking elicits accurate decoding in diffusion language models},
  author={Lee, Sanghyun and Kim, Seungryong and Park, Jongho and Park, Dongmin},
  journal={arXiv preprint arXiv:2511.05563},
  year={2025}
}

@article{du2026r,
  title={{R$^2$-dLLM}: Accelerating Diffusion Large Language Models via Spatio-Temporal Redundancy Reduction},
  author={Du, Zhenbang and Xia, Kejing and Zhong, Xinrui and Fu, Yonggan and Oswald, Nicolai and Ji, Binfei and Khailany, Brucek and Molchanov, Pavlo and Lin, Yingyan},
  journal={arXiv preprint arXiv:2604.18995},
  year={2026}
}

@article{fang2026locally,
  title={Locally confident, globally stuck: The quality-exploration dilemma in diffusion language models},
  author={Fang, Liancheng and Liu, Aiwei and Zou, Henry Peng and Chen, Yankai and Ma, Enze and Pan, Leyi and Miao, Chunyu and Huang, Wei-Chieh and Liu, Xue and Yu, Philip S},
  journal={arXiv preprint arXiv:2604.00375},
  year={2026}
}

@article{zhai2026core,
  title={CoRe: Context-Robust Remasking for Diffusion Language Models},
  author={Zhai, Kevin and Mollah, Sabbir and Wang, Zhenyi and Shah, Mubarak},
  journal={arXiv preprint arXiv:2602.04096},
  year={2026}
}

@article{cai2026confidence,
  title={Confidence-Based Decoding is Provably Efficient for Diffusion Language Models},
  author={Cai, Changxiao and Li, Gen},
  journal={arXiv preprint arXiv:2603.22248},
  year={2026}
}

@article{chen2025beyond,
  title={Beyond Confidence: Adaptive and Coherent Decoding for Diffusion Language Models},
  author={Chen, Kecheng and Liu, Ziru and Tao, Xijia and Liu, Hui and Fu, Xinyu and Zhang, Suiyun and Tu, Dandan and Kong, Lingpeng and Liu, Rui and Li, Haoliang},
  journal={arXiv preprint arXiv:2512.02044},
  year={2025}
}

@article{li2025diffusion,
  title={Diffusion language models know the answer before decoding},
  author={Li, Pengxiang and Zhou, Yefan and Muhtar, Dilxat and Yin, Lu and Yan, Shilin and Shen, Li and Vosoughi, Soroush and Liu, Shiwei},
  journal={arXiv preprint arXiv:2508.19982},
  year={2025}
}

@article{zhao2026context,
  title={Context Tokens are Anchors: Understanding the Repetition Curse in dMLLMs from an Information Flow Perspective},
  author={Zhao, Qiyan and Zhang, Xiaofeng and Chang, Shuochen and Chen, Qianyu and Yuan, Xiaosong and Chen, Xuhang and Liu, Luoqi and Zhang, Jiajun and Zhang, Xu-Yao and Wang, Da-Han},
  journal={arXiv preprint arXiv:2601.20520},
  year={2026}
}

@misc{huang2026empiricalanalysisdecodingbiases,
      title={Empirical Analysis of Decoding Biases in Masked Diffusion Models}, 
      author={Pengcheng Huang and Tianming Liu and Zhenghao Liu and Yukun Yan and Shuo Wang and Tong Xiao and Zulong Chen and Maosong Sun},
      year={2026},
      eprint={2508.13021},
      archivePrefix={arXiv},
      primaryClass={cs.AI},
      url={https://arxiv.org/abs/2508.13021}, 
}

@article{hong2026mitigating,
  title={Mitigating Mask Prior Drift and Positional Attention Collapse in Large Diffusion Vision-Language Models},
  author={Hong, Sujung and Yoon, Chanyong and Hwang, Seongjae},
  journal={arXiv preprint arXiv:2605.14530},
  year={2026}
}

@article{you2025llada,
  title={Llada-v: Large language diffusion models with visual instruction tuning},
  author={You, Zebin and Nie, Shen and Zhang, Xiaolu and Hu, Jun and Zhou, Jun and Lu, Zhiwu and Wen, Ji-Rong and Li, Chongxuan},
  journal={arXiv preprint arXiv:2505.16933},
  year={2025}
}

@article{li2026lavida,
  title={Lavida: A large diffusion language model for multimodal understanding},
  author={Li, Shufan and Kallidromitis, Konstantinos and Bansal, Hritik and Gokul, Akash and Kato, Yusuke and Kozuka, Kazuki and Kuen, Jason and Lin, Zhe and Chang, Kai-Wei and Grover, Aditya},
  journal={Advances in Neural Information Processing Systems},
  volume={38},
  pages={105101--105134},
  year={2026}
}

@article{hamilton2025lost,
  title={Lost in Space: Optimizing Tokens for Grammar-Constrained Decoding},
  author={Hamilton, Sil and Mimno, David},
  journal={arXiv e-prints},
  pages={arXiv--2502},
  year={2025}
}

@inproceedings{ghazvininejad2019mask,
  title={Mask-predict: Parallel decoding of conditional masked language models},
  author={Ghazvininejad, Marjan and Levy, Omer and Liu, Yinhan and Zettlemoyer, Luke},
  booktitle={Proceedings of the 2019 conference on empirical methods in natural language processing and the 9th international joint conference on natural language processing (EMNLP-IJCNLP)},
  pages={6112--6121},
  year={2019}
}

@article{austin2021structured,
  title={Structured denoising diffusion models in discrete state-spaces},
  author={Austin, Jacob and Johnson, Daniel D and Ho, Jonathan and Tarlow, Daniel and Van Den Berg, Rianne},
  journal={Advances in neural information processing systems},
  volume={34},
  pages={17981--17993},
  year={2021}
}

@article{lu2025adablock,
  title={Adablock-dllm: Semantic-aware diffusion llm inference via adaptive block size},
  author={Lu, Guanxi and Chen, Hao Mark and Karashima, Yuto and Wang, Zhican and Fujiki, Daichi and Fan, Hongxiang},
  journal={arXiv preprint arXiv:2509.26432},
  year={2025}
}

@article{zhang1,
  title={From Redundancy to Relevance: Enhancing Explainability in Multimodal Large Language Models},
  author={Zhang, Xiaofeng and Shen, Chen and Yuan, Xiaosong and Yan, Shaotian and Xie, Liang and Wang, Wenxiao and Gu, Chaochen and Tang, Hao and Ye, Jieping},
  journal={NAACL},
  year={2024}
}

@inproceedings{
zhang2,
title={Shallow Focus, Deep Fixes: Enhancing Shallow Layers Vision Attention Sinks to Alleviate Hallucination in {LVLM}s},
author={Xiaofeng Zhang and Yihao Quan and Chen Shen and Chaochen Gu and Xiaosong Yuan and Shaotian Yan and Jiawei Cao and Hao Cheng and Kaijie Wu and Jieping Ye},
booktitle={Knowledgeable Foundation Models at ACL 2025},
year={2025},
url={https://openreview.net/forum?id=B0buEm1XNH}
}

@article{zhang3,
  title={Simignore: Exploring and enhancing multimodal large model complex reasoning via similarity computation},
  author={Zhang, Xiaofeng and Zeng, Fanshuo and Gu, Chaochen},
  journal={Neural Networks},
  pages={107059},
  year={2024}
}

@article{zhang4,
  title={Enhancing Multimodal Large Language Models Complex Reason via Similarity Computation},
  author={Zhang, Xiaofeng and Zeng, Fanshuo and Quan, Yihao and Hui, Zheng and Yao, Jiawei},
  journal={Proceedings of the AAAI Conference on Artificial Intelligence},
  year={2025}
}

@article{zhang5,
  title={What drives attention sinks? A study of massive activations and rotational positional encoding in large vision--language models},
  author={Zhang, Xiaofeng and Zhu, Yuanchao and Gu, Chaochen and Cao, Jiawei and Cheng, Hao and Wu, Kaijie},
  journal={Information Processing \& Management},
  volume={63},
  number={2},
  pages={104431},
  year={2026},
  publisher={Elsevier}
}

@inproceedings{
zhang6,
title={Context Tokens are Anchors: Understanding the Repetition Curse in dMLLMs from an Information Flow Perspective},
  author={Zhao, Qiyan and Zhang, Xiaofeng and Chang, Shuochen and Chen, Qianyu and Yuan, Xiaosong and Chen, Xuhang and Liu, Luoqi and Zhang, Jiajun and Zhang, Xu-Yao and Wang, Da-Han},
booktitle={The Fourteenth International Conference on Learning Representations},
year={2026},
}

@inproceedings{
zhang7,
  title={Hallucination Begins Where Saliency Drops},
  author={Zhang, Xiaofeng and Zhu, Yuanchao and Gu, Chaochen and Yuan, Xiaosong and Zhao, Qiyan and Cao, Jiawei and Tang, Feilong and Fan, Sinan and Shen, Yaomin and Shen, Chen and others},
booktitle={The Fourteenth International Conference on Learning Representations},
year={2026},
}

@article{zhang8,
  title={D3ToM: Decider-Guided Dynamic Token Merging for Accelerating Diffusion MLLMs},
  author={Chang, Shuochen and Zhang, Xiaofeng and Liu, Qingyang and Niu, Li},
  journal={arXiv preprint arXiv:2511.12280},
  year={2025}
}

@article{zhang9,
  title={Diffusion-CAM: Faithful Visual Explanations for dMLLMs},
  author={Zuo, Haomin and Li, Yidi and Yang, Luoxiao and Zhang, Xiaofeng},
  journal={arXiv preprint arXiv:2604.11005},
  year={2026}
}

\appendix

\section{Appendix Overview}
\label{app:overview}
\begin{table*}[t]
\centering
\small
\setlength{\tabcolsep}{3.6pt}
\renewcommand{\arraystretch}{1.13}
\caption{Ablation study of IGFD components across three dMLLMs.}
\label{tab:ablation_components}
\resizebox{\textwidth}{!}{
\begin{tabular}{llcccccccc}
\toprule
\multirow{2}{*}{Model} & \multirow{2}{*}{Variant}
& \multicolumn{4}{c}{LLaVA-Bench}
& \multicolumn{3}{c}{CHAIR}
& \multicolumn{1}{c}{MathVista} \\
\cmidrule(lr){3-6}
\cmidrule(lr){7-9}
\cmidrule(lr){10-10}
& & all$\uparrow$ & conv$\uparrow$ & detail$\uparrow$ & complex$\uparrow$
& $C_S\downarrow$ & $C_i\downarrow$ & recall$\uparrow$
& acc$\uparrow$ \\
\midrule

\multirow{5}{*}{LLaDA-V}
& Confidence
& 50.4 & 43.4 & 57.5 & 63.6
& 5.6 & 4.0 & 36.6
& 30.8 \\
& w/o neighborhood need
& 71.7 & 61.9 & 73.1 & 79.1
& 5.3 & 3.8 & 37.0
& 33.5 \\
& w/o structural penalty
& 72.1 & 62.4 & 73.8 & 79.7
& 5.4 & 3.9 & 36.9
& 33.7 \\
& w/o dynamic frontier
& 72.0 & 62.2 & 73.5 & 79.5
& 5.4 & 3.9 & 37.0
& 33.6 \\
\rowcolor{blue!10}
& \textbf{Full IGFD}
& \textbf{72.8} & \textbf{63.1} & \textbf{74.6} & \textbf{80.8}
& \textbf{5.2} & \textbf{3.7} & \textbf{37.1}
& \textbf{34.2} \\
\midrule

\multirow{5}{*}{MMaDA}
& Confidence
& 35.1 & 34.4 & 39.7 & 33.7
& 24.6 & 11.3 & 39.5
& 23.5 \\
& w/o neighborhood need
& 45.9 & 40.1 & 49.2 & 41.9
& 20.1 & 10.4 & 40.1
& 24.8 \\
& w/o structural penalty
& 46.5 & 40.7 & 49.8 & 42.8
& 19.8 & 10.2 & 40.3
& 25.1 \\
& w/o dynamic frontier
& 46.2 & 40.5 & 49.6 & 42.4
& 19.6 & 10.0 & 40.4
& 25.0 \\
\rowcolor{blue!10}
& \textbf{Full IGFD}
& \textbf{47.3} & \textbf{41.9} & \textbf{50.6} & \textbf{43.7}
& \textbf{18.7} & \textbf{9.8} & \textbf{40.6}
& \textbf{25.9} \\
\midrule

\multirow{5}{*}{LaViDa}
& Confidence
& 51.1 & 32.8 & 57.2 & 60.1
& 6.8 & 9.8 & 34.8
& 41.6 \\
& w/o neighborhood need
& 77.2 & 63.6 & 77.4 & 77.9
& 5.5 & 5.8 & 35.5
& 44.8 \\
& w/o structural penalty
& 77.8 & 64.2 & 77.9 & 78.6
& 5.8 & 5.9 & 35.7
& 45.0 \\
& w/o dynamic frontier
& 77.6 & 64.0 & 77.7 & 78.4
& 5.7 & 5.9 & 35.7
& 44.9 \\
\rowcolor{blue!10}
& \textbf{Full IGFD}
& \textbf{78.7} & \textbf{65.1} & \textbf{78.6} & \textbf{79.7}
& \textbf{5.1} & \textbf{5.7} & \textbf{36.1}
& \textbf{45.3} \\

\bottomrule
\end{tabular}
}
\end{table*}
This appendix provides additional details to support the reproducibility and interpretation of IGFD.
We include implementation details, baseline configurations, detailed ablation studies, efficiency analysis, hyperparameter sensitivity analysis, and limitations.
These materials clarify the experimental protocol and provide a more complete understanding of how each component contributes to the final decoding performance.

\section{Implementation Details}
\label{app:implementation}

All experiments are conducted under the same model configuration for each backbone, and the only difference among compared methods lies in the decoding strategy.
We evaluate IGFD on three diffusion-based multimodal large language models: LLaDA-V, MMaDA, and LaViDa.
For all methods, we use deterministic decoding with temperature set to $0$.
At each decoding step, the model predicts the token distribution for all currently masked positions, and the selected positions are committed using the top-1 prediction.

IGFD uses the following commitment score:
\begin{equation}
s_{t,i} = \alpha \cdot \mathrm{conf}_{t,i}
+ \beta \cdot \mathrm{ig}_{t,i}
- \gamma \cdot \mathrm{struct}(\hat{x}_{t,i}),
\end{equation}
where $\mathrm{conf}_{t,i}$ is the token confidence, $\mathrm{ig}_{t,i}$ is the confidence-weighted neighborhood utility term, and $\mathrm{struct}(\hat{x}_{t,i})$ indicates whether the predicted token is a structural token.
Unless otherwise specified, we set $\alpha=0.7$, $\beta=0.5$, $\gamma=0.2$, and neighborhood radius $r=2$.

The neighborhood utility term is computed as:
\begin{equation}
\mathrm{ig}_{t,i} = \mathrm{conf}_{t,i} \cdot \mathrm{need}_{t,i},
\end{equation}
where $\mathrm{need}_{t,i}$ is the average entropy of masked neighboring positions within radius $r$.
This design encourages IGFD to prioritize tokens that are both reliable themselves and useful for reducing uncertainty in nearby unresolved positions.

For structural token detection, we use a lightweight tokenizer-level rule.
A token is treated as structural if it belongs to punctuation symbols, whitespace-only tokens, newline or tab markers, end-of-sequence tokens, or tokenizer-specific formatting markers.
This rule does not require additional models, parsers, or external linguistic tools.
The broader token categories used in the commitment behavior analysis, such as content tokens, relation tokens, function words, punctuation, and formatting tokens, are used only for visualization and analysis rather than for the core decoding algorithm.
Unless otherwise specified, we set the neighborhood radius to $r=2$, the frontier expansion radius to $R=2$, and the maximum frontier size to $F=8$.
The total decoding budget $T$ follows the default setting of each backbone model and is kept identical across all compared methods.
For a generation length of $N$, the per-step commitment budget is computed as 
$k_t=\lfloor N/T \rfloor + \mathbb{I}[t \leq N \bmod T]$.
Thus, IGFD primarily changes the commitment order and introduces only lightweight additional computation, without increasing the number of denoising steps, generation length, or model forward evaluations.

\section{Baseline Configurations}
\label{app:baseline_config}

\begin{figure*}[t]
    \centering
    \includegraphics[width=\textwidth]{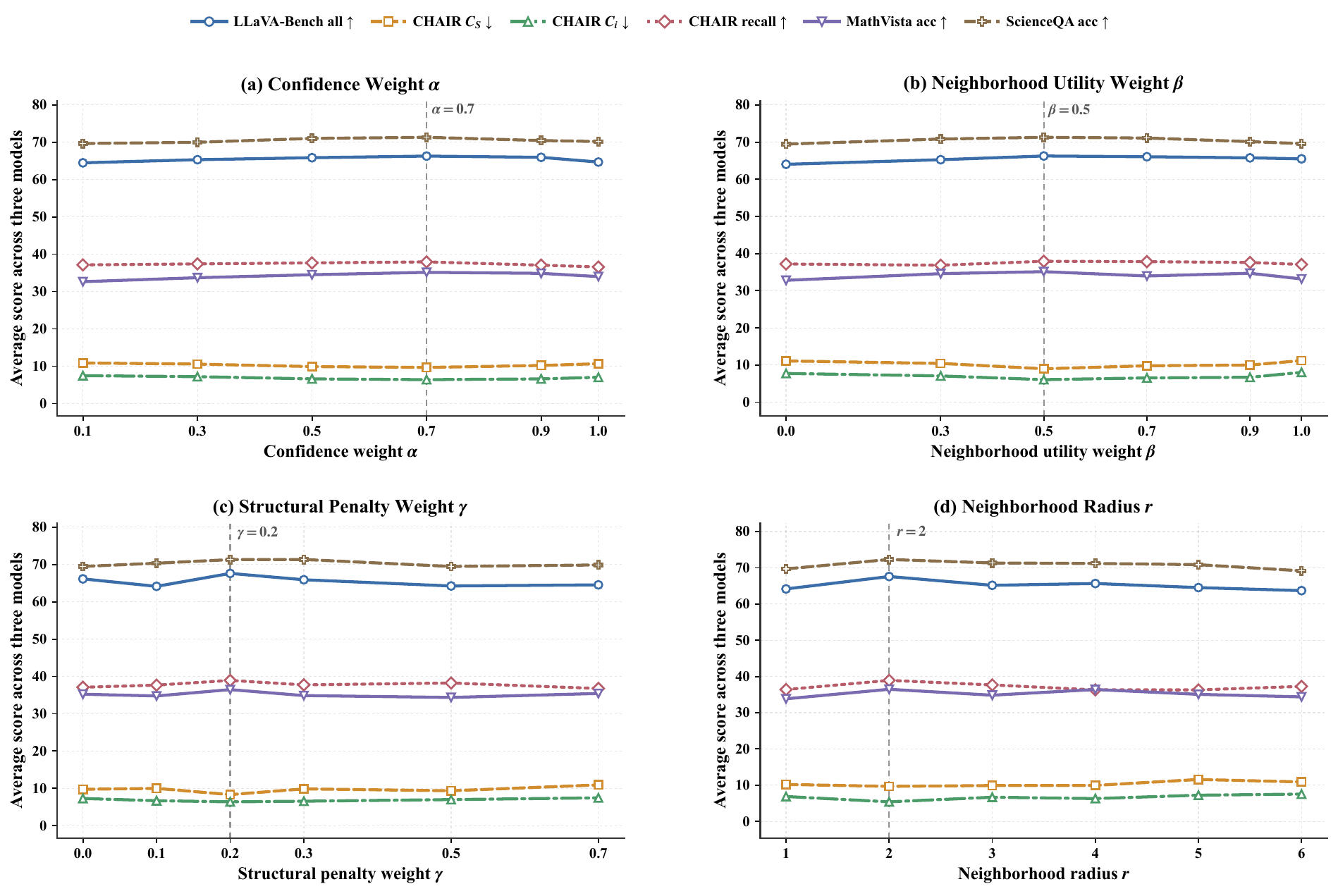}
    \caption{
    Hyperparameter sensitivity analysis of the proposed decoding strategy.
    We evaluate four key hyperparameters:
    (a) the confidence weight $\alpha$,
    (b) the neighborhood utility weight $\beta$,
    (c) the structural penalty weight $\gamma$,
    and (d) the neighborhood radius $r$.
    Each curve reports the average performance across three diffusion-based multimodal language models, including LLaDA-V, MMaDA, and LaViDa.
    The results show that moderate values of $\alpha$, $\beta$, and $\gamma$ generally achieve better trade-offs across perception, reasoning, and hallucination-related metrics, while an overly large structural penalty or neighborhood radius may degrade performance.
    The dashed vertical line in each subfigure indicates the selected default setting used in our main experiments.
    }
    \label{fig:hyperparameter_sensitivity}
\end{figure*}
To ensure a fair comparison, all methods are evaluated with the same backbone model, decoding budget, prompt format, and deterministic decoding setting. 
The compared methods differ only in how they select masked positions for commitment. 
Original decoding commits tokens mainly according to confidence over the global masked sequence. 
AdaBlock introduces adaptive block-level scheduling, while Wavefront restricts commitment to a progressively expanding frontier region. 
In contrast, IGFD uses the same decoding budget and does not introduce additional training, auxiliary models, or extra forward passes, but changes the commitment criterion from confidence-only ranking to an information-guided score that combines token reliability, neighborhood utility, and structural risk.

This comparison highlights that IGFD is designed as a decoding-time ranking strategy rather than a model modification. 
Therefore, the observed performance differences mainly come from the commitment order selected by each decoding strategy under the same computational budget.

\section{Details of Ablation Studies}
\label{app:ablation_details}

We provide detailed per-model ablation results to further analyze the contribution of each component in IGFD.
The \textit{Confidence} variant uses confidence-based commitment.
The \textit{w/o structural penalty} variant removes the structural risk term by setting $\gamma=0$.
The \textit{w/o neighborhood need} variant removes the neighborhood utility term and relies mainly on token confidence and structural penalty.
The full IGFD variant uses all components.

The results show that neighborhood utility, structural penalty, and dynamic frontier all contribute to the final performance.
Removing neighborhood need consistently lowers LLaVA-Bench and MathVista scores, indicating that uncertainty-aware local utility helps select more informative commitments.
Removing the structural penalty also degrades the results, especially on hallucination-related metrics, suggesting that delaying premature punctuation and formatting commitments is beneficial.
The w/o dynamic frontier variant selects candidates without the adaptive frontier constraint and performs worse than full IGFD, showing that dynamic frontier expansion helps maintain sufficient local context during commitment.
Full IGFD achieves the strongest overall performance across the three backbones.

These results are consistent across all evaluated backbones, suggesting that the effectiveness of IGFD does not depend on a specific diffusion architecture.
In particular, the gains are simultaneously reflected in perception quality, reasoning accuracy, and hallucination reduction, indicating that the proposed commitment strategy improves generation quality from multiple aspects rather than optimizing for a single metric.




\section{Hyperparameter Sensitivity}
\label{app:hyperparameter}







We further analyze the sensitivity of IGFD to the confidence weight $\alpha$, information-guided utility weight $\beta$, structural penalty weight $\gamma$, and neighborhood radius $r$.
The default setting is $\alpha=0.7$, $\beta=0.5$, $\gamma=0.2$, and $r=2$.

As shown in Figure~\ref{fig:hyperparameter_sensitivity}, IGFD is relatively stable around the default setting.
A moderate confidence weight is necessary to ensure that committed tokens remain reliable.
The neighborhood utility weight improves the ability to prioritize tokens that are useful for nearby uncertain positions, but an overly large value may overemphasize uncertain regions.
Similarly, the structural penalty helps delay premature punctuation and formatting tokens, while too large a penalty may excessively postpone necessary structural decisions.
The best performance is obtained around $r=2$, suggesting that local neighboring masks provide the most useful uncertainty signal for commitment ordering.
Overall, the relatively smooth performance variations across different settings suggest that IGFD is not overly sensitive to hyperparameter choices.
This robustness makes the method easier to apply across different diffusion-based multimodal language models without extensive tuning.

\end{document}